\pdfoutput=1

\documentclass[11pt]{article}

\usepackage{emnlp2021}
\usepackage{times}
\usepackage{latexsym}
\usepackage{graphicx}
\usepackage{multirow}
\usepackage{booktabs}
\usepackage{comment}
\usepackage{caption}
\usepackage{amsmath}
\usepackage{subcaption}
\usepackage{natbib}
\usepackage{mathtools}
\usepackage{amssymb, bm}
\usepackage{mathrsfs}
\usepackage{CJKutf8}
\usepackage{float}
\graphicspath{ {./sections/figs/} }

\newcommand*{\Eb}{\mathbb{E}}

\DeclarePairedDelimiterXPP{\KL}[2]{D_\text{KL}}\lbrack\rbrack{}{{#1} \delimsize\Vert {#2}}
\DeclarePairedDelimiterXPP{\Prob}[1]{\Pb}\lbrace\rbrace{}{#1}
\DeclarePairedDelimiterXPP{\Ev}[1]{\Eb}\lbrack\rbrack{}{#1}
\DeclarePairedDelimiterXPP{\Evr}[2]{\Eb_{#1}}\lbrack\rbrack{}{#2}

\DeclarePairedDelimiter{\abs}{\lvert}{\rvert}

\newcommand*{\vect}[1]{\bm{#1}}

\usepackage[T1]{fontenc}

\usepackage[utf8]{inputenc}

\usepackage{microtype}

\newcommand{\veryshortarrow}[1][3pt]{\mathrel{%
   \hbox{\rule[\dimexpr\fontdimen22\textfont2-.2pt\relax]{#1}{.4pt}}%
   \mkern-4mu\hbox{\usefont{U}{lasy}{m}{n}\symbol{41}}}}

\title{Looking for Clues of Language in Multilingual BERT to Improve Cross-lingual Generalization}

\author{Chi-Liang Liu\thanks{$^*$Equal Contribution} \quad Tsung-Yuan Hsu$^*$ \quad Yung-Sung Chuang \quad Chung-Yi Li \quad Hung-Yi Lee\\
College of Electrical Engineering and Computer Science, National Taiwan University \\
{ \tt \{liangtaiwan1230, sivia89024,  tlkagkb93901106\}@gmail.com} \\
{ \tt\{b05901033, r07942080\}@ntu.edu.tw}\\
}

\date{}

\hypersetup{final}  
\begin{document}
\maketitle
\begin{abstract}
Token embeddings in multilingual BERT (m-BERT) contain both language and semantic information. 
We find that the representation of a language can be obtained by simply averaging the embeddings of the tokens of the language. 
Given this language representation, we control the output languages of multilingual BERT by manipulating the token embeddings, thus achieving unsupervised token translation. 
We further propose a computationally cheap but effective approach
to improve the cross-lingual ability of m-BERT based on this observation.   
\end{abstract}

\section{Introduction}
Multilingual BERT (m-BERT)~\cite{devlin:19} has demonstrated its strength in cross-lingual transfer on a variety of tasks~\citep{conneau:18, wu:19, hsu-etal:19, pires:19}; this has been credited to the cross-lingual alignment of its internal representations, in which semantically similar or functionally similar words from different languages are represented with similar embeddings~\cite{Cao2020Multilingual,why_mBERT}. 

How is language information embedded in m-BERT?
The answer may be more straightforward than expected.
We find that the averaged token embedding of a language well represents the language.
To verify this observation, we show that if an English sentence is input to m-BERT, and its embeddings are then shifted in a specific direction in the embedding space, m-BERT outputs a sentence in another language semantically close to the input sentence. 

After showing the existence of language information in the embeddings of m-BERT and having an easy way to extract it, we eliminate these language-specific variations in the embeddings, and demonstrate that this is a practical way to boost the zero-shot cross-lingual transferability of m-BERT on downstream tasks.

In the literature, some have attempted to improve the cross-lingual alignment of pre-trained m-BERT. For example, \citet{Cao2020Multilingual} propose finetuning m-BERT on small parallel dataset, and \citet{libovick2020language} propose zero-centering the embedding language by language to achieve language neutrality and demonstrate progress on retrieval tasks, all under an unsupervised scenario. Our work is concurrent to \citep{libovick2020language, greek}'s and bears similarities to its approaches. We discussed more applications of language representation compared to them.

The contributions of this work can be summarized as the following:
\begin{itemize}
  \item Language information in m-BERT can be represented by the average of
  all token embeddings of the specific language. This is verified by
  unsupervised token translation.
  \item The cross-lingual transferability of m-BERT in downstream tasks can be
  improved by manipulating token embeddings.
\end{itemize}

\section{Language Representation}
\label{section:methods}

We assume that we have $n$ languages denoted by $\{L_1, L_2, \ldots, L_n\}$
and their corresponding corpora. 

\paragraph{Context-Dependent Representation}
Given an input sequence~$x$ and token index~$i$, we denote the hidden
representation in layer~$l$ by $\bm{h}^l_{x_i}$.

\paragraph{Language Mean} 
Given a language~$L$ and its corresponding corpus~$C$ composed of a set of
sentences~$x$, we denote the language mean of layer~$l$ by
\[\vect{R}^l_{L}=\mathop{\Eb}_{x_i \in C}\left[\vect{h}^l_{x_i}\right],\]
which represents the mean of all the token embeddings in the corpora.
We assume that the language mean contains language-specific information but no
semantic information.

Although this assumption about language representation here seems naive, in the
experiments, we show that $\vect{R}^l_{L}$ well represents the language information
in token embeddings.
For each language~$L$, there is a language-specific representation
$\vect{R}^l_{L}$ for each layer~$l$.
Because we do not know for which layer~$l$ $\vect{R}^l_{L}$ best represents
language~$L$, $l$ is a hyperparameter in the following algorithms.

\paragraph{Zero-mean} 
To eliminate language-specific information, 
we simply subtract language mean $\vect{R}^l_{L_k}$ from each token 
embedding $\vect{h}^l_{x_i}$, thus
moving the token embedding to a language-agnostic joint space.
The language-agnostic hidden representation $\hat{\vect{h}}^l_{x_i}$ can be
written as
\begin{equation}
    \hat{\vect{h}}^l_{x_i} = \vect{h}^l_{x_i} - \vect{R}^l_{L_k},
    \label{eq:zero}
\end{equation}
where $\vect{h}^l_{x_i}$ is extracted from token~$x_i$ in $L_k$.

\paragraph{Mean Difference Shift (MDS)}
\begin{CJK*}{UTF8}{gbsn}
In addition to eliminating language information, we can move the embedding in the
space of $L_1$ to the space of $L_2$: this amounts to unsupervised token
translation.
That is, given the embedding of \textit{me} in English, we modify the
embedding to cause it to be interpreted by m-BERT as the embedding of 我{}
(\textit{me} in Chinese).
\end{CJK*}

We feed a sentence~$L_1$ into m-BERT and extract the embedding of each token at
layer~$l$. 
Then we subtract $\vect{R}^l_{L_1}$ from the embedding as in (\ref{eq:zero}) to
remove the information of $L_1$, and then add $\vect{R}^l_{L_2}$  to shift
the embedding to the $L_2$ space.
Formally, we modify token embedding $\vect{h}^l_{x_i}$ in $L_1$ into embedding
$\tilde{\vect{h}}^l_{x_i}$ in $L_2$ as
\begin{equation}
    \tilde{\vect{h}}^l_{x_i} = \vect{h}^l_{x_i} + \vect{R}^l_{L_2}-\vect{R}^l_{L_1}.
    \label{eq:MDS}
\end{equation}







\section{Unsupervised Token Translation} 
In this section, we show that the implicit language-specific information in
the embedding space can be disentangled from semantic embeddings. 
We use MDS to input a sentence in $L_1$ to m-BERT and
\textit{translate} it to a sentence in $L_2$.

\subsection{Setup}

The formulation of MDS is a slight modification of (\ref{eq:MDS}):
$\tilde{\vect{h}}^l_{x_i} = \vect{h}^l_{x_i} +
\alpha\left(\vect{R}^l_{L_2}-\vect{R}^l_{L_1}\right)$, where $\alpha$ is a
hyperparameter; we will see its influence in the experimental
results.
Given the input, the token embeddings are modified at a specific layer $l$. 
The $(l+1)$-th layer takes the modified embeddings as input, and the final layer
generates a sequence of tokens. 
The sentences in this experiment are from XNLI test-set, which contains 15
languages, including low-resource languages such as Swahili and Urdu. 

\subsection{Evaluation metrics}

We use two different metrics to analyze the results of unsupervised token
translation quantitatively. 
\paragraph{BLEU-1 Score} This metric measures the translation quality without
considering the fluency of the converted sequence.

\paragraph{Conversion Rate}
Besides translation quality, we also calculated the \emph{conversion rate}: the percentage of
tokens converted from the source language to the target language, which is defined as
\[ \text{conversion rate} = \frac{\text{\# of } y \in (V_t-V_s)}{\text{\# of } y - \text{\# of } y \in V_s\cap V_t},\]
where $y$ is output tokens of the model and $V_s$ and $V_t$ are the token sets of the
source and target language. 
As tokens shared by both vocabularies are not taken into account, they are
excluded from the numerator and denominator terms.

\subsection{Results}
Surprisingly, we were able to produce the predicted tokens in language $L_2$ given $L_1$
input by applying MDS; many of the predicted tokens were the token-level
translation of the input tokens in $L_1$, even for low-resource languages.
Sample output is shown in Appendix~\ref{sec:appendix}.

\begin{table*}[ht!]
\centering
\caption{Quantitative unsupervised token translation results using 10th BERT layer}
\label{tab:mt}
\footnotesize
\setlength\tabcolsep{3pt}
\begin{tabular}{l|cccccc|cccccc}
\toprule
& en$\veryshortarrow$de & en$\veryshortarrow$fr & en$\veryshortarrow$ur & en$\veryshortarrow$sw & en$\veryshortarrow$zh & en$\veryshortarrow$el & de$\veryshortarrow$en & fr$\veryshortarrow$en & ur$\veryshortarrow$en & sw$\veryshortarrow$en & zh$\veryshortarrow$en & el$\veryshortarrow$en \\
\midrule

BLEU-1 ($\alpha$=1) & 7.53  & 8.53  & 5.56  & 7.96  & 15.25 & 7.88  & 7.34  & 9.08  & 5.52  & 6.34  & 4.37  & 6.54   \\
BLEU-1 ($\alpha$=2) &  8.03  & 10.24 & 6.31  & 7.23  & 21.51 & 14.91 & 7.48  & 8.52  & 6.23  & 7.48  & 5.38  & 6.65   \\
BLEU-1 ($\alpha$=3) & 12.35 & 10.65 & 5.35  & 7.16  & 15.95 & 19.13 & 6.29  & 12.27 & 5.74  & 6.45  & 6.17  & 4.73   \\

\midrule\midrule
Conversion rate ($\alpha$=1) & 40.2 & 41.7 & 61.1 & 15.3 & 47.8 & 62.1 & 45.2 & 49.6 & 29.9 & 14.7 & 23.9 & 30.2\\
Conversion rate ($\alpha$=2) & 74.8 & 75.7 & 99.4 & 97.4 & 90.0 & 99.1 & 67.3 & 601 & 83.0 & 65.6 & 60.8 & 97.9\\
Conversion rate ($\alpha$=3) & 95.2 & 96.3 & 99.8 & 100 & 99.5 & 100 & 79.5 & 73.1 & 96.6 & 93.6 & 91.4 & 99.7\\
\bottomrule
\end{tabular}
\vspace{-0.5cm}
\end{table*}


Table~\ref{tab:mt} shows the quantitative results. First, although the
translation result falls short of existing unsupervised
translation methods~\citep{kim2018improving}, it constitutes strong evidence that we
can use MDS to manipulate language-specific information in the token embedding space
and then induce m-BERT to switch from one language to another. 
Second, we observe that as $\alpha$ increases, the model converts more
tokens to target language $L_2$ and never decodes tokens that do not belong to both $L_1$
and $L_2$. Given a negative $\alpha$, the model always decodes tokens
belonging to $L_1$. 
This shows that in the embedding space, the direction related
to language is unique. 
We offer a further analysis of $\alpha$ in Appendix~\ref{sec:appendix}.

\begin{table}[t!]
    \caption{Tatoeba sentence retrieval using 8th BERT layer}
    \label{tab:tatoeba}
    \centering
    \footnotesize
    \setlength\tabcolsep{5pt}
    \begin{tabular}{l|cccccc}
    \toprule
    Method &de	&es	&ar	&el	&fr	&hi	\\ \midrule
    Original   &75.4	&64.1	&24.5	&29.8	&64.3	&\bf{34.8}	\\
    MUSE  &1.3	&0.2	&0.3	&0.5	&23.8	&0.2	\\
    Zero-mean  &73.5	&61.8	&23.5	&29.4	&63.7	&29.9	\\
    MDS &\bf{76.8}	&\bf{67.5}	&\bf{29.1}	&\bf{30.6}	&\bf{67.0}	&31.4\\
    \midrule
    &ru	&vi &th	&tr	&zh \\ \midrule
    Original &\bf{63.6}	&\bf{61.0}	&13.7	&32.9	&68.6 \\
    MUSE 	&0.2	&0.2	&0.4	&0.1	&0.2 \\
    Zero-mean 	&59.6	&51.2	&13.7	&32.8	&64.1 \\
    MDS &59.4	&51.5	&\bf{17.5}	&\bf{36.8}	&\bf{69.2}   \\
	\bottomrule
    \end{tabular}
    \vspace{-0.25cm}
\end{table}

\begin{table}[t!]
    \caption{BUCC2018 dev set and test set sentence retrieval using 8th BERT layer}
    \label{tab:bucc2018}
    \centering
    \footnotesize
    \setlength\tabcolsep{5pt}
    \begin{tabular}{l|l|cccc}
    \toprule
	&Method &de & fr & ru & zh \\ \midrule
    &Original &75.62	&72.07	&68.59	&66.04 \\
    Dev &Zero-mean &71.10	&70.51	&65.92	&59.91\\
    &MUSE &13.68	&57.11	&40.49	&16.98\\
    &MDS&\bf{76.91}	&\bf{73.45}	&\bf{71.60}	&\bf{66.91}\\
    
    \midrule
    &Original &63.22 	&62.47	&11.65  &50.47\\
    Test &Zero-mean&59.59   &59.25  &10.40  &45.42\\
    &MUSE &59.00	&11.30	&6.03	&2.03\\
    &MDS    &\bf{65.76}   &\bf{63.95} &\bf{12.36}  &\bf{52.45}\\
    
	\bottomrule
    \end{tabular}
    \vspace{-0.5cm}
\end{table}

\section{Cross-lingual Sentence Retrieval}
Extracting parallel sentences from a comparable corpus between two languages is
a common way to evaluate cross-lingual
embeddings~\citep{hu2020xtreme, zweigenbaum-etal-2017-overview,
artetxe2018massively}. 
In this section, we use evaluations on a sentence-level retrieval task to 
demonstrate that MDS achieve better cross-lingually aligned.

\subsection{Task}
We evaluate the effect of MDS and zero-mean on two sentence retrieval tasks:
BUCC2018 and Tatoeba.
We use the mean vector of all token embeddings in a sentence as the sentence
embedding and cosine similarity as the distance metric. 
Token embeddings were extracted from a specific layer of the BERT encoder, and MDS
or zero-mean shifts pre-computed on the whole dataset were applied directly to
the extracted embeddings. 

\subsection{MDS vs Zero-Mean}
\label{sec:mds_zm_diff}

Although applying MDS or zero-mean in a sentence retrieval task seem
similar at first glance, there are subtle differences. 
Assume sentence embeddings $\vect{v}_1 \in L_1$ and $\vect{v}_2 \in
L_2$, and that these two sentences in different languages have the same semantic
meaning.
Assume there exist \textit{real} language representations $\vect{R}_{L_1}^*$
and $\vect{R}_{L_2}^*$ that perfectly eliminate language from the
embedding\footnote{Unknown to us} such that
$\vect{v}_1-\vect{R}_{L_1}^*=\vect{v}_2-\vect{R}_{L_2}^*$.
Language representations $\vect{R}_{L_1}$ and $\vect{R}_{L_2}$ obtained via
averaging are approximations of the real representations,\footnote{Superscript $^l$ 
ignored here for simplicity} and $\delta_1$ and $\delta_2$ are the differences
between the real and approximate language representations. 
\[
\begin{split}
	 &\vect{v}_1-\vect{R}_{L_1} \neq \vect{v}_2-\vect{R}_{L_2} \\
  \rightarrow& \vect{v}_1-\vect{R}_{L_1}-\vect{\delta}_1=\vect{v}_2-\vect{R}_{L_2}-\vect{\delta}_2
\end{split}
\]
Then the post-MDS and post-zero-mean cosine similarities of $\vect{v}_1$ and $\vect{v}_2$ 
are 
\[
\small
\begin{split}
\cos_\text{MDS}\left(\vect{v}_1,\vect{v}_2\right)
&= \frac{ \left(\vect{v}_1-{\vect{R_{L_1}}}+\vect{R_{L_2}}\right) \cdot \vect{v}_2}
{\left\lvert\vect{v}_1-{\vect{R_{L_1}}}+\vect{R_{L_2}}\right\rvert\left\lvert\vect{v}_2\right\rvert}\\
&= \frac{ \left(\vect{v}_2+\vect{\delta}\right) \cdot \vect{v_2}}
{\left\lvert \vect{v}_2+\vect{\delta}\right\rvert\left\lvert\vect{v}_2\right\rvert}, \text{ where } \vect{\delta}=\vect{\delta}_1-\vect{\delta}_2\\\
\cos_\text{zero-mean}\left(\vect{v}_1,\vect{v}_2\right)
&= \frac{ \left(\vect{v}_1-{\vect{R_{L_1}}}\right) \cdot \left(\vect{v}_2 -{\vect{R_{L_2}} }\right)}
{\left\lvert\vect{v}_1-{\vect{R_{L_1}}}\right\rvert\left\lvert\vect{v}_2-{\vect{R_{L_2}}}\right\rvert}\\
&= \frac{
\left(\vect{v}_2- \vect{R^*_{L_2}} + \vect{\delta}_1\right) \left(\vect{v}_2- \vect{R^*_{L_2}} + \vect{\delta}_2\right)}
{\left\lvert\vect{v}_2- \vect{R^*_{L_2}} + \vect{\delta}_1\right\rvert \left\lvert\vect{v}_2- \vect{R^*_{L_2}} + \vect{\delta}_2\right\rvert}.\\
\end{split}
\]
This shows that zero-mean is more sensitive to 
approximation error when $\left\lvert\vect{v}_2 \right\rvert >
\left\lvert\vect{v}_2- \vect{R_{L_2}}\right\rvert >
\max\left(\abs{\vect{\delta_1},\abs{\vect{\delta_2}},\abs{\vect{\delta}}}\right)$.\footnote{This
is very possible. Because $\vect{v}_2$ is in $L_2$, it may have the same
direction as $\vect{R}_{L_2}$.}
The differences between the two methods are further verified in the experiments.



\subsection{Results}
\begin{table*}[t!]
\centering
\caption{POS tagging results}
\label{tab:pos}
\footnotesize
\begin{tabular}{l|ccccccccccccc|c}
    \toprule
  Method  & ar   & bg   & de   & el   & es   & fr   & hi   & ru   & th   & tr   & ur   & vi   & zh   & Average  \\
    \midrule
 Original    & 53.8 & 85.4 & 86.2 & 81.1 & 86.1 & 42.9 & 66.8 & 85.5 & 41.7 & 68.6 & 56.3 & \textbf{53.8} & 61.8 & 66.9      \\
Zero-mean & \textbf{54.3} & 86.1 & \textbf{86.6} & \textbf{81.8} & 86.6 & 43.7 & 68.1 & \textbf{86.5} & 41.6 & \textbf{69.7} & 56.6 & 53.4 & 62.5 & 67.5     \\
MDS & 54.2 & \textbf{86.4} & 86.5 & 81.5 & \textbf{86.8} & \textbf{43.9} & \textbf{68.9} & 86.4 & \textbf{44.2} & 69.4 & \textbf{57.1} & 52.4 & \textbf{63.0}   & \textbf{67.8}     \\
    \bottomrule
\end{tabular}
\end{table*}


\begin{table*}[ht!]
	 \caption{Dependency parsing results. Numbers are Labeled Attachment Score(LAS).}
    \label{tab:dep}
    \centering
    \footnotesize
    \begin{tabular}{l|ccccccccccccc|c}
    \toprule
Method & ar & bg & de & el & es & fr & hi & ru & th & tr & ur & vi & zh & Average  \\
\midrule
Original & 28.2 & 70.7 & \textbf{74.0} & \textbf{71.6} & 72.1 & 74.8 & 35.3 & 69.0 & 30.8 & 32.9 & 28.3 & \textbf{37.8} & \textbf{35.4} & 50.8 \\
Zero-mean & \textbf{28.2} & \textbf{71.0} & 73.4 & 71.4 & 72.2 & \textbf{75.7} & 36.3 & \textbf{69.3} & \textbf{32.5} & \textbf{34.6} & 28.6 & 37.0 & 35.2 & \textbf{51.2} \\
MDS & 28.0 & 70.8 & 73.7 & 71.1 & \textbf{72.2} & 75.3 & \textbf{36.5} & 68.8 & 30.4 & 34.2 & \textbf{29.0} & 35.6 & 35.0 & 50.8 \\
	\bottomrule
    \end{tabular}
    \vspace{-0.5cm}
\end{table*}

The sentence retrieval results are shown in Tables~\ref{tab:bucc2018} and
\ref{tab:tatoeba}. 
On the BUCC2018 dev set and test set, MDS-shifted embeddings consistently
yield higher accuracies on all languages, and zero-meaned embeddings
are worse than doing nothing. 
On the Tatoeba test set, the MDS embeddings
are also the best in most languages,
except for Hindi, Russian, and Vietnamese. We also tried using rotation matrices to align the embedding (MUSE, \cite{lample2017unsupervised}), but found that the unsupervised alignment method is not working on BERT.


\section{Cross-Lingual Transfer}

\subsection{Setup}
In zero-shot cross-lingual transfer learning, m-BERT was fine-tuned on the
source language, which was English in the following experiments, and tested on
languages never seen during fine-tuning.
For each language, we used around 5M tokens from Wikipedia documents to compute
the language representations.

\paragraph{Zero-mean} 
During fine-tuning, we applied zero-mean on the token embeddings of the source
language and forwarded the modified embeddings to the remaining layers during
training. 
During testing, we fed the fine-tuned model with target language data and
applied zero-mean to the embeddings at layer~$l$ as well. 
The language vector means were extracted from Wikipedia data using the
pre-trained model.

\paragraph{MDS} 
In this approach, we did not modify embeddings during training. 
During testing, we applied MDS to the embeddings at layer~$l$. 
The mean difference vectors were extracted from Wikipedia data using
the fine-tuned model.

\subsection{Tasks}
To show that the proposed methods improve cross-lingual zero-shot learning performance,
we conducted experiments on two tasks: part-of-speech (POS) tagging and
dependency parsing.

\paragraph{Part-of-Speech Tagging} For POS tagging, we used the Universal
Dependencies v2.5~\citep{udpos} treebanks for 
90 languages. Each word was assigned one of 17 universal
POS tags. 
The model was trained on English and tested on 13 other languages. 

\paragraph{Dependency Parsing} For dependency parsing, the dataset and cross-lingual transfer settings were exactly the same as POS tagging.

\subsection{Results}

Tables~\ref{tab:pos} and \ref{tab:dep} compare the results of the baselines
and our methods on POS tagging and dependency parsing, respectively. 
For POS tagging, zero-mean and MDS both improve the performance on
the testing sets across languages, with only a few exceptions.
MDS was not helpful in Thai (th), and neither approach improved on
Vietnamese (vi).
For dependency parsing, contrary to the previous observations, we found that only zero-mean improved upon the baselines, while MDS didn't. Since more linguistic factors affects dependency parsing (e.g. head-directionality parameter) than POS tagging, we felt that more analyses are needed to explain the performance of dependency parsing.

\section{Conclusion}
In this paper, we examine the existence of language-specific information in
m-BERT embeddings and achieve unsupervised token translation
by manipulating language-specific information. 
The proposed methods are further shown to be effective in improving
cross-lingual embedding alignment and cross-lingual transfer learning.
We will further explore the proposed approach on more downstream tasks.



\bibliography{anthology,emnlp2020}

\begin{thebibliography}{15}
\expandafter\ifx\csname natexlab\endcsname\relax\def\natexlab#1{#1}\fi

\bibitem[{Artetxe and Schwenk(2018)}]{artetxe2018massively}
Mikel Artetxe and Holger Schwenk. 2018.
\newblock \href {http://arxiv.org/abs/1812.10464} {Massively multilingual
  sentence embeddings for zero-shot cross-lingual transfer and beyond}.

\bibitem[{Cao et~al.(2020)Cao, Kitaev, and Klein}]{Cao2020Multilingual}
Steven Cao, Nikita Kitaev, and Dan Klein. 2020.
\newblock \href {https://openreview.net/forum?id=r1xCMyBtPS} {Multilingual
  alignment of contextual word representations}.
\newblock In \emph{International Conference on Learning Representations}.

\bibitem[{Conneau et~al.(2018)Conneau, Rinott, Lample, Williams, Bowman,
  Schwenk, and Stoyanov}]{conneau:18}
Alexis Conneau, Ruty Rinott, Guillaume Lample, Adina Williams, Samuel Bowman,
  Holger Schwenk, and Veselin Stoyanov. 2018.
\newblock \href {https://doi.org/10.18653/v1/D18-1269} {{XNLI}: Evaluating
  cross-lingual sentence representations}.
\newblock In \emph{Proceedings of the 2018 Conference on Empirical Methods in
  Natural Language Processing}, pages 2475--2485, Brussels, Belgium.
  Association for Computational Linguistics.

\bibitem[{Devlin et~al.(2019)Devlin, Chang, Lee, and Toutanova}]{devlin:19}
Jacob Devlin, Ming-Wei Chang, Kenton Lee, and Kristina Toutanova. 2019.
\newblock \href {https://doi.org/10.18653/v1/N19-1423} {{BERT}: Pre-training of
  deep bidirectional transformers for language understanding}.
\newblock In \emph{Proceedings of the 2019 Conference of the North {A}merican
  Chapter of the Association for Computational Linguistics: Human Language
  Technologies, Volume 1 (Long and Short Papers)}, pages 4171--4186,
  Minneapolis, Minnesota. Association for Computational Linguistics.

\bibitem[{Gonen et~al.(2020)Gonen, Ravfogel, Elazar, and Goldberg}]{greek}
Hila Gonen, Shauli Ravfogel, Yanai Elazar, and Yoav Goldberg. 2020.
\newblock \href {https://doi.org/10.18653/v1/2020.blackboxnlp-1.5} {It{'}s not
  {G}reek to m{BERT}: Inducing word-level translations from multilingual
  {BERT}}.
\newblock In \emph{Proceedings of the Third BlackboxNLP Workshop on Analyzing
  and Interpreting Neural Networks for NLP}, pages 45--56, Online. Association
  for Computational Linguistics.

\bibitem[{Hsu et~al.(2019)Hsu, Liu, and Lee}]{hsu-etal:19}
Tsung-Yuan Hsu, Chi-Liang Liu, and Hung-yi Lee. 2019.
\newblock \href {https://doi.org/10.18653/v1/D19-1607} {Zero-shot reading
  comprehension by cross-lingual transfer learning with multi-lingual language
  representation model}.
\newblock In \emph{Proceedings of the 2019 Conference on Empirical Methods in
  Natural Language Processing and the 9th International Joint Conference on
  Natural Language Processing (EMNLP-IJCNLP)}, pages 5933--5940, Hong Kong,
  China. Association for Computational Linguistics.

\bibitem[{Hu et~al.(2020)Hu, Ruder, Siddhant, Neubig, Firat, and
  Johnson}]{hu2020xtreme}
Junjie Hu, Sebastian Ruder, Aditya Siddhant, Graham Neubig, Orhan Firat, and
  Melvin Johnson. 2020.
\newblock \href {http://arxiv.org/abs/2003.11080} {Xtreme: A massively
  multilingual multi-task benchmark for evaluating cross-lingual
  generalization}.

\bibitem[{Kim et~al.(2018)Kim, Geng, and Ney}]{kim2018improving}
Yunsu Kim, Jiahui Geng, and Hermann Ney. 2018.
\newblock Improving unsupervised word-by-word translation with language model
  and denoising autoencoder.
\newblock In \emph{Proceedings of the 2018 Conference on Empirical Methods in
  Natural Language Processing}, pages 862--868.

\bibitem[{Lample et~al.(2017)Lample, Conneau, Denoyer, and
  Ranzato}]{lample2017unsupervised}
Guillaume Lample, Alexis Conneau, Ludovic Denoyer, and Marc'Aurelio Ranzato.
  2017.
\newblock Unsupervised machine translation using monolingual corpora only.
\newblock \emph{arXiv preprint arXiv:1711.00043}.

\bibitem[{Libovický et~al.(2020)Libovický, Rosa, and
  Fraser}]{libovick2020language}
Jindřich Libovický, Rudolf Rosa, and Alexander Fraser. 2020.
\newblock \href {http://arxiv.org/abs/2004.05160} {On the language neutrality
  of pre-trained multilingual representations}.

\bibitem[{Liu et~al.(2020)Liu, Hsu, Chuang, and yi~Lee}]{why_mBERT}
Chi-Liang Liu, Tsung-Yuan Hsu, Yung-Sung Chuang, and Hung yi~Lee. 2020.
\newblock What makes multilingual {BERT} multilingual?
\newblock In \emph{arXiv}.

\bibitem[{Nivre et~al.(2020)Nivre, de~Marneffe, Ginter, Hajic, Manning,
  Pyysalo, Schuster, Tyers, and Zeman}]{udpos}
Joakim Nivre, Marie-Catherine de~Marneffe, Filip Ginter, Jan Hajic,
  Christopher~D. Manning, Sampo Pyysalo, Sebastian Schuster, Francis Tyers, and
  Daniel Zeman. 2020.
\newblock \href {https://www.aclweb.org/anthology/2020.lrec-1.497} {Universal
  dependencies v2: An evergrowing multilingual treebank collection}.
\newblock In \emph{Proceedings of The 12th Language Resources and Evaluation
  Conference}, pages 4034--4043, Marseille, France. European Language Resources
  Association.

\bibitem[{Pires et~al.(2019)Pires, Schlinger, and Garrette}]{pires:19}
Telmo Pires, Eva Schlinger, and Dan Garrette. 2019.
\newblock \href {https://doi.org/10.18653/v1/P19-1493} {How multilingual is
  multilingual {BERT}?}
\newblock In \emph{Proceedings of the 57th Annual Meeting of the Association
  for Computational Linguistics}, pages 4996--5001, Florence, Italy.
  Association for Computational Linguistics.

\bibitem[{Wu and Dredze(2019)}]{wu:19}
Shijie Wu and Mark Dredze. 2019.
\newblock \href {https://doi.org/10.18653/v1/D19-1077} {Beto, bentz, becas: The
  surprising cross-lingual effectiveness of {BERT}}.
\newblock In \emph{Proceedings of the 2019 Conference on Empirical Methods in
  Natural Language Processing and the 9th International Joint Conference on
  Natural Language Processing (EMNLP-IJCNLP)}, pages 833--844, Hong Kong,
  China. Association for Computational Linguistics.

\bibitem[{Zweigenbaum et~al.(2017)Zweigenbaum, Sharoff, and
  Rapp}]{zweigenbaum-etal-2017-overview}
Pierre Zweigenbaum, Serge Sharoff, and Reinhard Rapp. 2017.
\newblock \href {https://doi.org/10.18653/v1/W17-2512} {Overview of the second
  {BUCC} shared task: Spotting parallel sentences in comparable corpora}.
\newblock In \emph{Proceedings of the 10th Workshop on Building and Using
  Comparable Corpora}, pages 60--67, Vancouver, Canada. Association for
  Computational Linguistics.

\end{thebibliography}
\bibliographystyle{acl_natbib}

\clearpage
\appendix

\section{Analysis of $\alpha$ in Unsupervised Token Translation}

\label{sec:appendix}
\begin{table}[h]
    \label{tab:common}
    \caption{Size of token set and size of English token set intersection with another language token set}
    \centering
    \footnotesize
    \setlength\tabcolsep{3pt}
    \begin{tabular}{c|ccccccc}
    \toprule
	& en & de & fr & el & zh & ur & sw \\ \midrule
	$|V_{\text{lang}}|$ & 9140 & 9212 & 8552 & 3189 & 3866 & 4085 & 5609 \\ \midrule
    $|V_{\mathit{en}}\cap V_{\text{lang}}|$ & 9140 & 3230 & 3911 & 1696 & 1325 & 1549 & 2970 \\
	\bottomrule
    \end{tabular}
    \vspace{-0.5cm}
\end{table}

\begin{figure}[ht]
    \vspace{-0.25cm}
    \begin{subfigure}[b]{\linewidth}
        \includegraphics[width=1.0\linewidth]{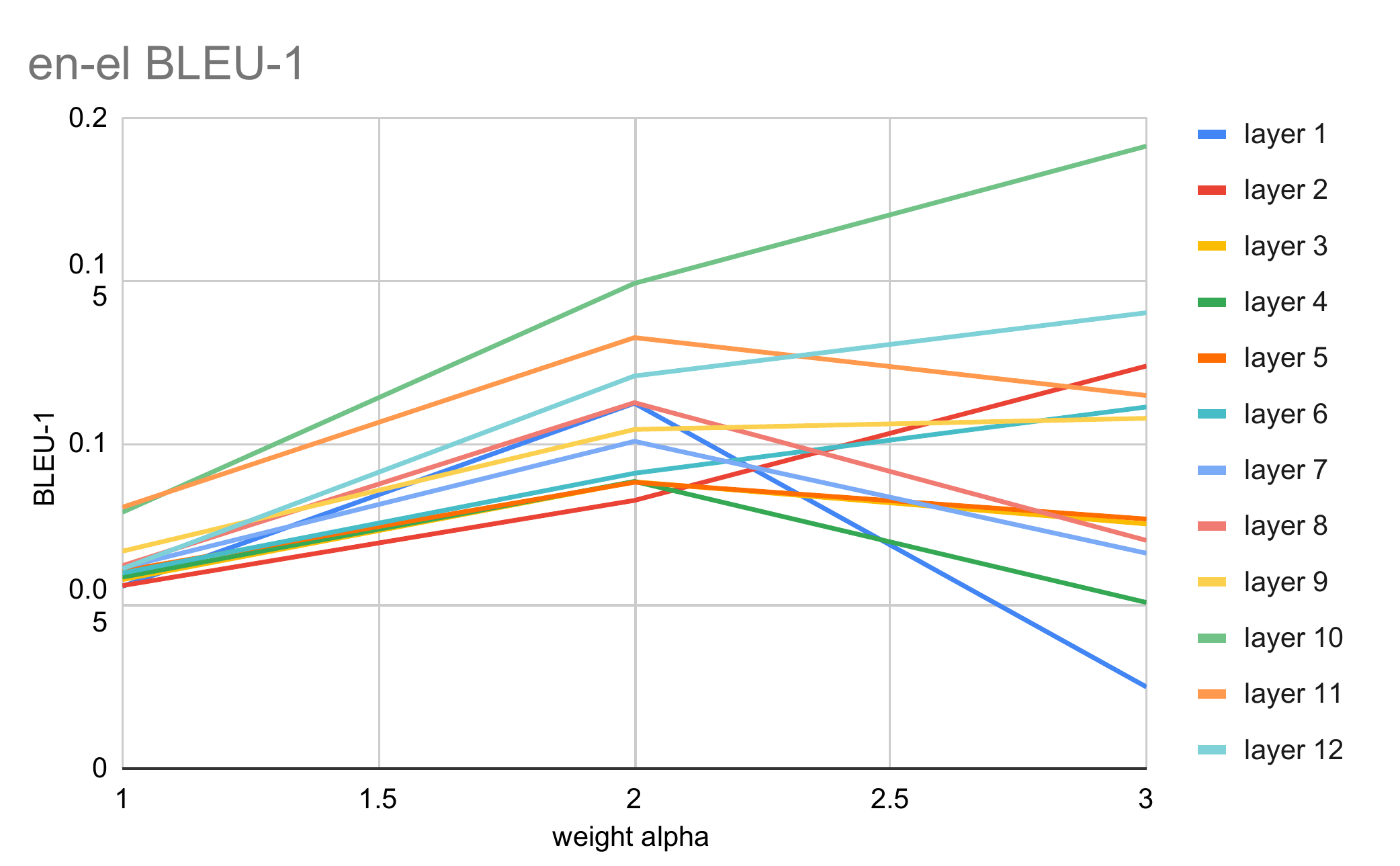}
        \subcaption{By $\alpha$}
    \end{subfigure}
    \begin{subfigure}[b]{\linewidth}
        \includegraphics[width=1.0\linewidth]{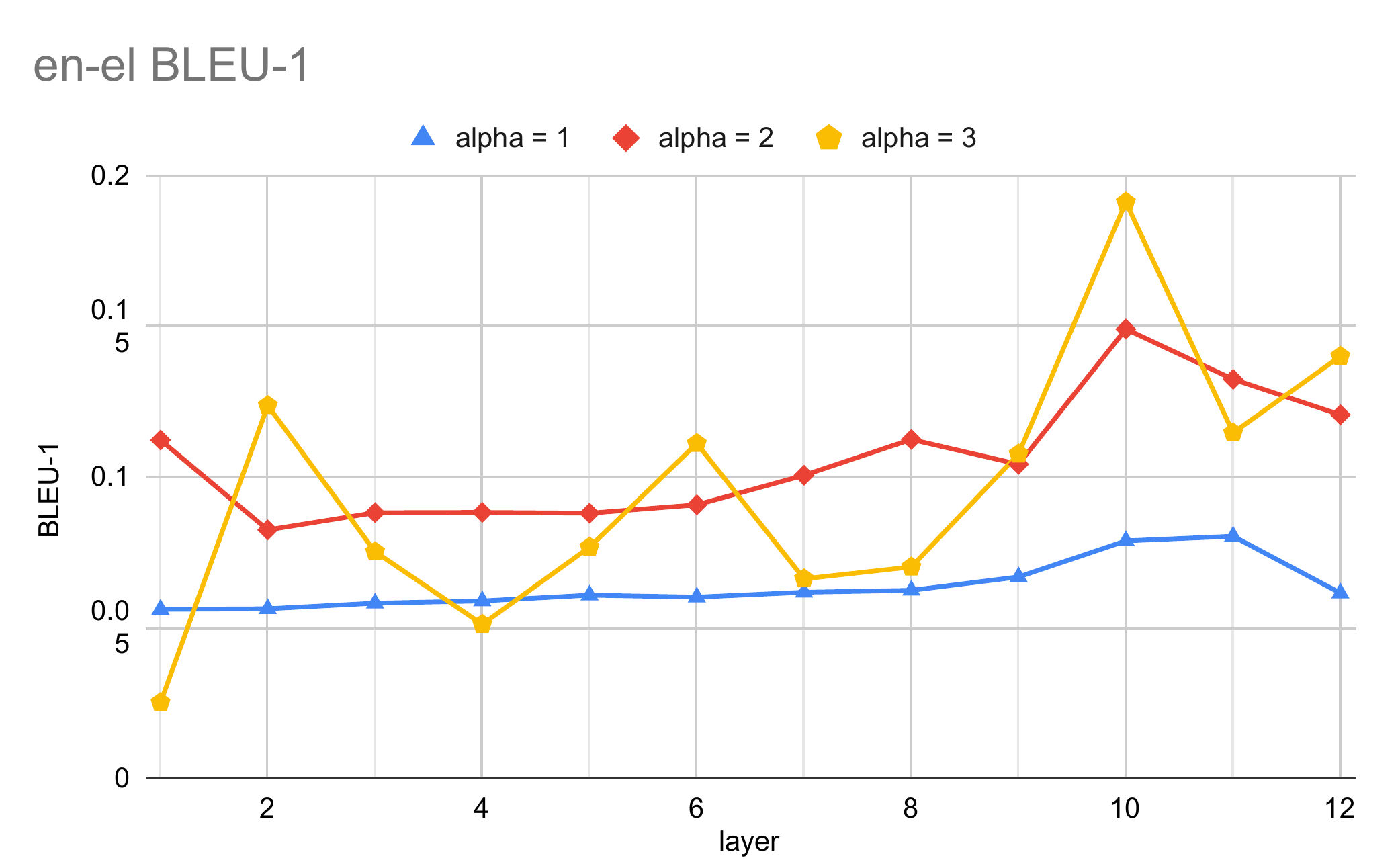}
        \subcaption{By layer}
    \end{subfigure}
	 \caption{Direction of change of BLEU-1 for unsupervised
	 \textit{en\,$\veryshortarrow$\,el} token translation for various $\alpha$ and
	 with different layers}
    \label{fig:bleuchange}
    \vspace{-0.25cm}
\end{figure} 


\begin{figure}[ht]
    \centering
    \includegraphics[width=1.0\linewidth]{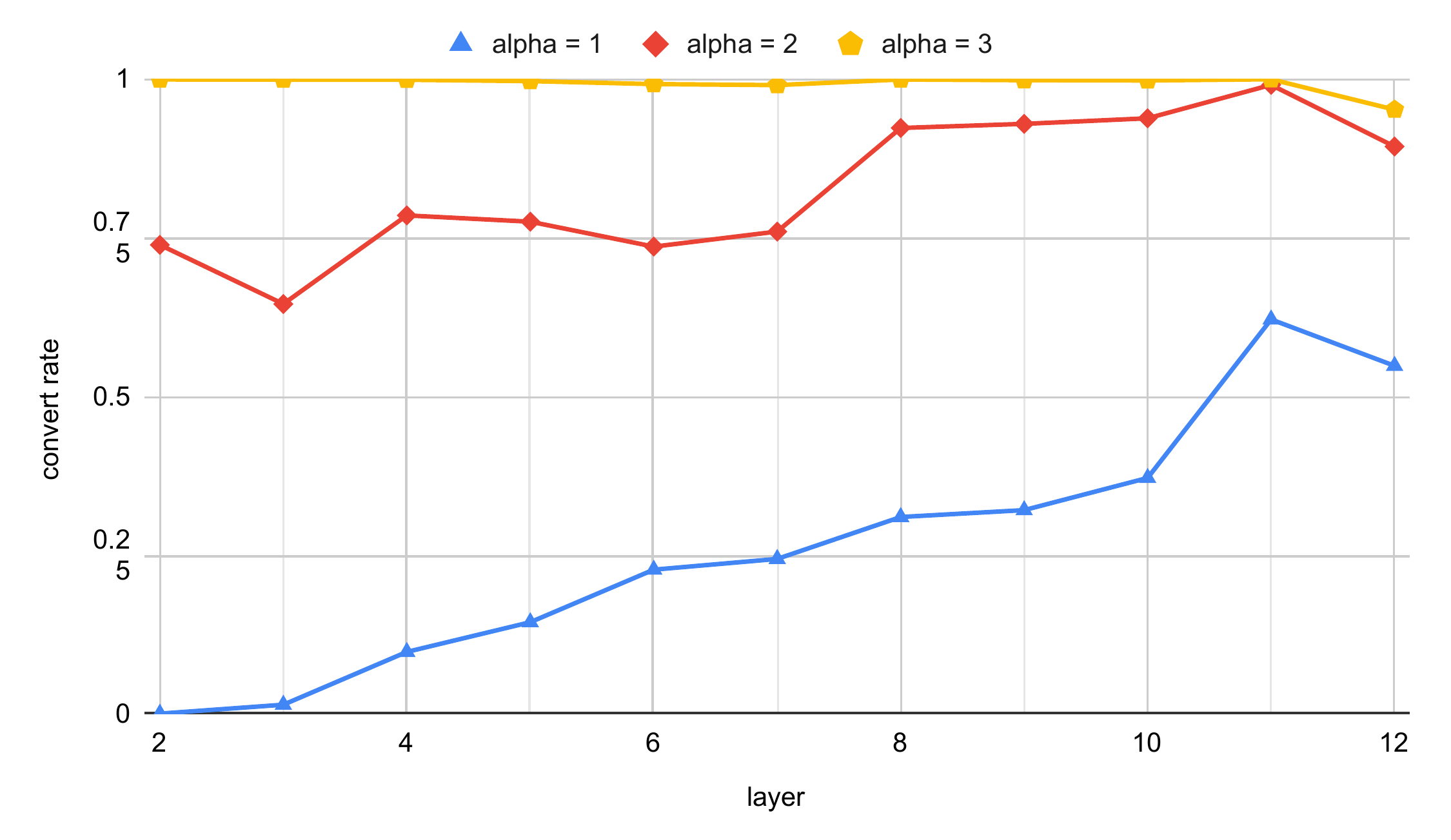}
	 \caption{Conversion rate on \textit{en\,$\veryshortarrow$\,el} data given
	 different $\alpha$ on different layers for MDS}
    \label{fig:convert}
\end{figure}
\begin{table*}[t!]
\centering
\caption{Unsupervised token translation of random sample (MDS, layer 10)}
\label{tab:ex}
\footnotesize
\setlength\tabcolsep{3pt}
\begin{CJK*}{UTF8}{gbsn}
\begin{tabular}{r|l}
\toprule
Input (en) & {The girl that can help me is all the way across town. There is no one who can help me.}\\
\midrule
Ground truth (zh) & 能帮助我的女孩在小镇的另一边。 没有人能帮助我。。 \\
en$\veryshortarrow$zh, $\alpha=1$ & . 孩 ， can 来 我 是 all the way across 市 。 。 There 是 无 人 人 can help 我 。\\
en$\veryshortarrow$zh, $\alpha=2$ & . 孩 的 的 家 我 是 这 个 人 的 市 。 。 他 是 他 人 人 的 到 我 。\\
en$\veryshortarrow$zh, $\alpha=3$ & 。 ， 的 的 的 他 是 的 个 的 的 ， 。 ： 他 是 他 人 ， 的 。 他 。\\
\midrule
Ground truth (fr) & La fille qui peut m'aider est à l'autre bout de la ville. Il n'y a personne qui pourrait m'aider.\\
en$\veryshortarrow$fr, $\alpha=1$ & . girl qui can help me est all la way across town . . There est no one qui can help me .\\
en$\veryshortarrow$fr, $\alpha=2$ & . girl qui de help me est all la way dans , . . Il est de seul qui pour aid me .\\
en$\veryshortarrow$fr, $\alpha=3$ & , , , de , me , all la , , , , , n n n n , , , , ,\\
\bottomrule
\end{tabular}
\end{CJK*}
\end{table*}

We present an example of \textit{en\,$\veryshortarrow$\,el} in
Figures~\ref{fig:bleuchange} and \ref{fig:convert} to show how the conversion rate
and BLEU-1 score change with different $\alpha$ weights and with different layers.

Despite the mixed influences of weight increases on BLEU-1, 
in the last few layers (10 or 11), the BLEU-1 of most
languages rose noticeably when $\alpha$ was set to $3.0$ (also shown in the best
layer row in Table~\ref{tab:mt}). 
This suggests that the last few layers are better for disentangling
language-specific representations, which is consistent with the observation in
the literature that the last few layers contain more language-specific
information for predicting masked words~\citep{pires:19}.

\end{document}